\documentclass[sigconf,screen]{acmart}
\setcopyright{none}
\renewcommand\footnotetextcopyrightpermission[1]{}
\acmConference[arXiv]{}{arXiv preprint}{2026}

\usepackage{multirow}
\usepackage{graphicx} 
\usepackage{booktabs} 
\usepackage{enumitem}
\usepackage[table]{xcolor}
\definecolor{mmpurple}{HTML}{64287E}
\definecolor{mygreen}{HTML}{00B050}
\definecolor{myblue}{HTML}{5b9bd5}
\definecolor{myorange}{HTML}{ff8000}
\definecolor{mygray}{HTML}{808080}
\definecolor{tbred}{HTML}{d62728}
\definecolor{tbblue}{HTML}{1f77b4}
\newcommand{\g}[1]{\textcolor{mygray}{#1}}
\newcommand{\tbredbf}[1]{\textcolor{tbred}{\textbf{#1}}}
\newcommand{\tbbluebf}[1]{\textcolor{tbblue}{\textbf{#1}}}
\begin{document}

\title{ProVG: Progressive Visual Grounding via Language Decoupling for Remote Sensing Imagery}

\settopmatter{authorsperrow=4}

\author{Ke Li}
\affiliation{%
  \institution{Xidian University}
  \country{China}}
\email{like0413@stu.xidian.edu.cn}

\author{Ting Wang}
\affiliation{%
  \institution{Xidian University}
  \country{China}}
\email{24031212234@stu.xidian.edu.cn}

\author{Di Wang}
\authornote{Corresponding author.}
\affiliation{%
  \institution{Xidian University}
  \country{China}}
\email{wangdi@xidian.edu.cn}

\author{Yongshan Zhu}
\affiliation{%
  \institution{Xidian University}
  \country{China}}
\email{zhu.ys@stu.xidian.edu.cn}

\author{Yiming Zhang}
\affiliation{%
  \institution{UC San Diego}
  \country{USA}}
\email{yiz134@ucsd.edu}

\author{Tao Lei}
\affiliation{%
  \institution{SUST}
  \country{China}}
\email{leitaoly@163.com}

\author{Quan Wang}
\affiliation{%
  \institution{Xidian University}
  \country{China}}
\email{qwang@xidian.edu.cn}

\begin{abstract}
Remote sensing visual grounding (RSVG) aims to localize objects in remote sensing imagery according to natural language expressions. Previous methods typically rely on sentence-level vision-language alignment, which struggles to exploit fine-grained linguistic cues, such as \textit{spatial relations} and \textit{object attributes}, that are crucial for distinguishing objects with similar characteristics. Importantly, these cues play distinct roles across different grounding stages and should be leveraged accordingly to provide more explicit guidance. In this work, we propose \textbf{ProVG}, a novel RSVG framework that improves localization accuracy by decoupling language expressions into global context, spatial relations, and object attributes. To integrate these linguistic cues, ProVG employs a simple yet effective progressive cross-modal modulator, which dynamically modulates visual attention through a \textit{survey-locate-verify} scheme, enabling coarse-to-fine vision-language alignment. In addition, ProVG incorporates a cross-scale fusion module to mitigate the large-scale variations in remote sensing imagery, along with a language-guided calibration decoder to refine cross-modal alignment during prediction. A unified multi-task head further enables ProVG to support both referring expression comprehension and segmentation tasks. Extensive experiments on two benchmarks, \textit{i.e.}, RRSIS-D and RISBench, demonstrate that ProVG consistently outperforms existing methods, achieving new state-of-the-art performance.
\end{abstract}

\maketitle

\begin{figure}[!t]
    \centering
    \includegraphics[width=\linewidth]{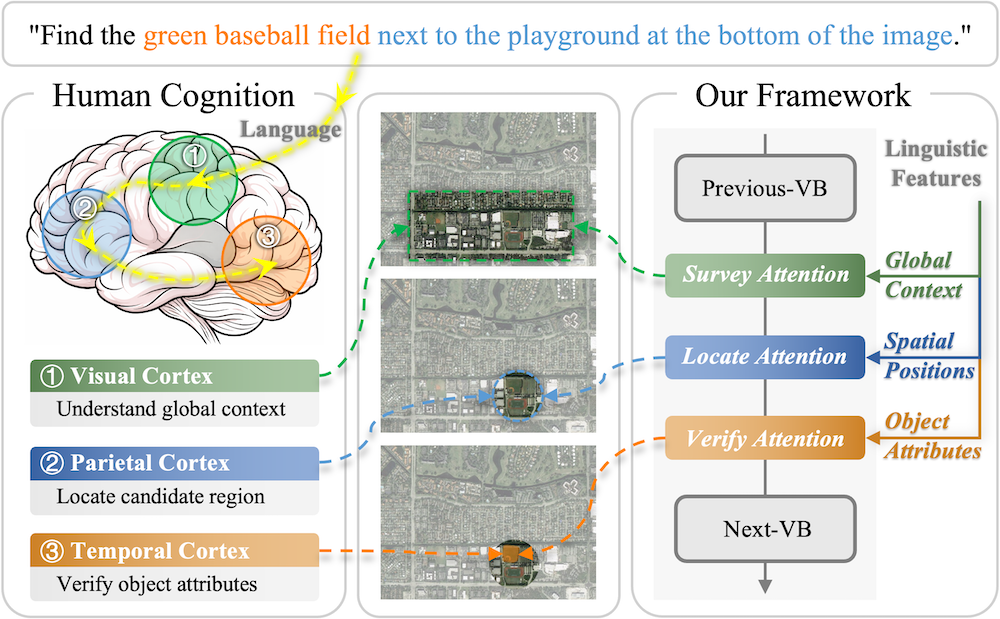}
    \caption{Left: Human cognition resolves referring expressions in a staged manner, involving global context understanding, spatial localization, and attribute verification. 
    Right: Inspired by this process, we propose a progressive grounding paradigm for RSVG, which decouples linguistic cues into global context, spatial relations, and object attributes, and integrates them via a \textit{survey-locate-verify} scheme to guide visual attention.}
    \label{fig:teaser}
\end{figure}

\section{Introduction}
\label{sec:intro}
Remote sensing visual grounding (RSVG) aims to localize a specific object in remote sensing imagery based on a natural language expression, encompassing both referring expression comprehension (RSREC)~\cite{sun2022visual,liu2025language,ding2025visual} and segmentation (RSRES)~\cite{li2025segearth_r1,li2026rsvg,yuan2023rrsis,yao2025remotesam}.
Unlike conventional remote sensing object detection and segmentation methods that operate on predefined categories, RSVG offers a more flexible and practical paradigm by allowing users to specify arbitrary targets via natural language.

Existing RSVG methods typically encode a language expression into a single sentence embedding as the reference.
For instance, MGVLF~\cite{zhan2023rsvg} and LQVG~\cite{lan2024language} employ cross-modal transformer decoders to fuse holistic sentence embeddings with image-level features for language comprehension.
However, simply combining the two modalities may result in ambiguous target identification, as objects in remote sensing imagery often exhibit diverse spatial scales and dense layouts.
To alleviate this issue, several works introduce language-guided modulation mechanisms into the visual backbone, \emph{e.g.}, LPVA~\cite{li2024language} and RMSIN~\cite{liu2024rotated}, which enhance expression-aware visual representations through carefully designed attention modulators.
Nevertheless, these methods still rely on sentence-level matching and fail to exploit fine-grained linguistic cues, particularly \emph{spatial relations} and \emph{object attributes}, that are critical for distinguishing visually similar targets.

Psycholinguistic and neuroscientific studies~\cite{tanenhaus1995integration,maunsell2006feature} suggest that human visual grounding follows a staged process: humans first build a global understanding of the scene, then refine search via spatial relations, and finally verify the target using object attributes.
These stages correspond to distinct cortical pathways involving the visual, parietal, and temporal cortices (Fig.~\ref{fig:teaser}). 
This observation implies a progressive grounding paradigm, in which linguistic cues are incorporated incrementally rather than jointly.

From this perspective, effective visual grounding should rely on structured reasoning over different types of linguistic cues rather than holistic interpretation of the expression. 
Referring expressions can naturally be decomposed into multiple components, among which \textbf{\textit{spatial relations}} and \textbf{\textit{object attributes}} play essential roles.
The former describes \emph{where} the target is located, \emph{e.g.}, absolute positions or relations to other objects, while the latter specifies \emph{what} the object looks like, \emph{e.g.}, category, appearance.
These observations raise a fundamental question: \textit{how can heterogeneous linguistic cues be effectively utilized for accurate visual grounding in remote sensing imagery?}

\begin{figure*}[!t]
    \centering
    \includegraphics[width=\linewidth]{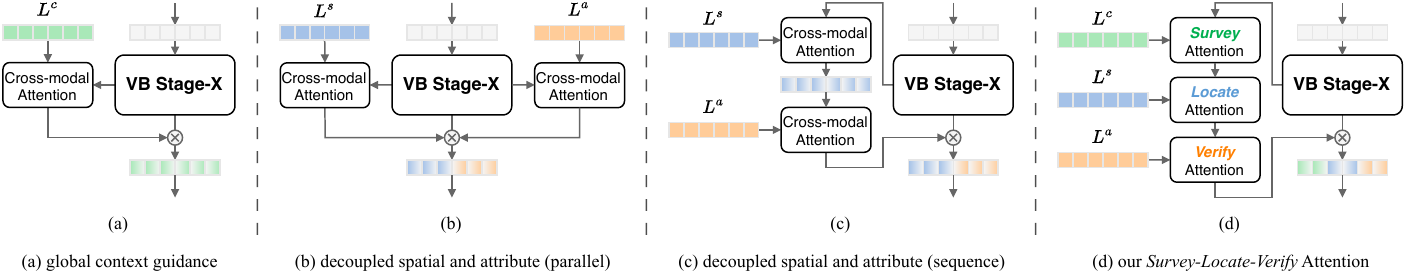}
    \caption{Architectural comparison of different cross-modal modulators. 
    (a) Global context guidance. 
    (b) Parallel injection of decoupled spatial and attribute cues. 
    (c) Sequential injection of decoupled spatial and attribute cues.
    (d) The proposed progressive cross-modal modulator based on \textit{survey-locate-verify} attention.
    Here, $L^c$, $L^s$, and $L^a$ denote the context, spatial, and attribute features, respectively, and VB-Stage-X represents the X-th stage of the visual backbone.}
    \label{fig:2}
\end{figure*}

To answer this question, we systematically investigate several strategies for incorporating linguistic cues into the visual backbone, including global context guidance (Fig.~\ref{fig:2}\textcolor{mmpurple}{(a)}), decoupled spatial and attribute cues injected in parallel (Fig.~\ref{fig:2}\textcolor{mmpurple}{(b)}) or sequentially (Fig.~\ref{fig:2}\textcolor{mmpurple}{(c)}), as well as a variant that combines global context with spatial and attribute cues (Fig.~\ref{fig:2}\textcolor{mmpurple}{(d)}).
Through a series of experiments (see Tab.~\ref{tab:modulator}), we observe that different linguistic cues provide complementary information, and combining global context, spatial relations, and object attributes leads to more accurate grounding. 
Moreover, sequential integration consistently outperforms parallel injection, suggesting that a structured, progressive grounding paradigm is more effective for RSVG.

Motivated by these findings, we propose a Progressive Cross-modal Modulator (PCM), which follows a \textit{survey-locate-verify} scheme to guide visual attention in a structured manner. 
PCM consists of three modules:
\emph{(i)} a survey attention module that leverages global context to establish an overall understanding of the scene, 
\emph{(ii)} a locate attention module that exploits spatial relations to focus on candidate regions, and
\emph{(iii)} a verify attention module that utilizes object attributes to confirm the target identity and resolve ambiguities. 
By explicitly decoupling linguistic cues and integrating them sequentially, PCM enables progressive vision-language alignment that mirrors the human staged perception mechanism.

Building upon this design, we present \textbf{ProVG}, a novel grounding framework for RSVG.
Specifically, PCM is embedded into multiple stages of the visual backbone to dynamically modulate multi-scale visual features.
A lightweight cross-scale fusion module (CFM) is introduced to effectively aggregate language-aware features across scales.
In addition, a language-guided calibration module (LCM) is integrated into the decoder to refine cross-modal alignment during prediction.
Finally, a unified multi-task prediction head is adopted to simultaneously support RSREC and RSRES tasks. 
Extensive experiments on RRSIS-D and RISBench datasets demonstrate the effectiveness of the proposed method.

Our contributions are summarized as follows:
\begin{itemize}[leftmargin=*]
    \item We provide an empirical perspective on RSVG by analyzing the decoupled structure of referring expressions, showing that progressive utilization of heterogeneous linguistic cues is crucial for accurate grounding.
    \item We propose a progressive cross-modal modulator based on a \textit{survey-locate-verify} scheme, which decouples heterogeneous linguistic cues and sequentially guides visual attention. Building upon this mechanism, we develop ProVG, a unified multi-task framework that achieves state-of-the-art performance on multiple benchmarks.
    \item Extensive experiments on two RSVG benchmarks demonstrate that ProVG outperforms existing methods and achieves significant gains in challenging scenarios.
\end{itemize}

\section{Related Work}
\label{sec2}
\subsection{Remote Sensing Visual Grounding}
\label{rw_rsvg}
RSVG aims to localize the referred object in remote sensing imagery given a natural language expression.
Compared with natural images, remote sensing imagery exhibits large scale variations, dense object distributions, and complex backgrounds, posing unique challenges for visual grounding~\cite{xiao2024towards}.
Existing RSVG studies are commonly evaluated under two settings: RSREC and RSRES.

\noindent \textbf{Remote Sensing Referring Expression Comprehension.}
RSREC focuses on localizing the referred object with a bounding box~\cite{hang2024regionally,wang2024multistage,zhao2024spatial,zhao2025context,li2025tacmt}.
Sun \textit{et al.}~\cite{sun2022visual} first introduce this task and propose GeoVG, which models geographic relations described in language using a geospatial relationship graph.
Inspired by DETR-like frameworks, subsequent studies adopt cross-modal transformer decoders that concatenate textual and visual features for joint decoding, such as MGVLF~\cite{zhan2023rsvg} and LQVG~\cite{lan2024language}.
However, such joint decoding schemes are limited in modeling complex vision-language interactions, particularly in remote sensing scenes with extreme scale variations and cluttered backgrounds, often leading to attention drift and suboptimal grounding.
To address this issue, subsequent works introduce language-aware attention mechanisms to modulate visual features at early stages of the visual backbone, including adaptive feature selection~\cite{ding2024visual} and cross-modal modulation strategies~\cite{li2024language,li2024injecting}.

\noindent \textbf{Remote Sensing Referring Expression Segmentation.}
Compared with RSREC, RSRES requires predicting a pixel-level mask for the referred object, demanding more precise cross-modal alignment~\cite{yao2026remotereasoner,xin2025segearth,zhan2025does}.
Most existing methods extend general referring expression segmentation frameworks, particularly LAVT~\cite{yang2022lavt}, which adopts a Swin Transformer~\cite{liu2021swin} visual backbone and a BERT~\cite{devlin2018bert} language encoder.
Yuan \textit{et al.}~\cite{yuan2023rrsis} first introduce the RSRES benchmark RegSegRS and design a language-guided cross-scale enhancement module to refine multi-scale visual features under linguistic guidance.
Subsequent studies improve segmentation performance by incorporating orientation-aware modeling~\cite{liu2024rotated} or designing stronger bidirectional cross-modal interaction mechanisms~\cite{dong2024cross}.
Recent works explore large language model (LLM) driven grounding frameworks.
For example, GeoGround~\cite{zhou2024geoground} formulates pixel-level grounding as a language-driven reasoning process by serializing grounding cues into textual sequences, while SegEarth-R1~\cite{li2025segearth_r1} integrates instruction-following language models with a segmentation pipeline for geospatial reasoning.

\noindent \textbf{Discussion.}
Despite these advances, existing RSVG methods fundamentally rely on holistic sentence-level representations, failing to capture the fine-grained linguistic cues that are essential for accurate grounding.
In contrast, the proposed ProVG explicitly decouples heterogeneous linguistic cues and progressively injects them through staged cross-modal interactions, enabling fine-grained vision-language alignment for precise grounding in challenging remote sensing scenarios.

\subsection{Multi-task Visual Grounding}
\label{rw_multitask}
Multi-task learning has been widely explored for jointly addressing REC and RES.
Early work by Luo \textit{et al.}~\cite{luo2020multi} proposes a collaborative multi-task network with a consistency energy maximization loss to align feature activations between the two tasks.
With the emergence of transformer architectures, subsequent studies adopt unified vision--language transformers with task-specific heads for joint prediction.
For example, RefTR~\cite{li2021referring} directly predicts bounding boxes and segmentation masks using decoder-generated contextualized language queries, while Polyformer~\cite{liu2023polyformer} further improves segmentation precision through floating-point coordinate modeling and multi-polygon generation.
Although transformer-based multi-task grounding models achieve promising performance, they often suffer from high computational costs.
To address this, EEVG~\cite{chen2024efficient} performs vision-language fusion in a transformer decoder and prunes background visual tokens based on attention scores.
Another line of work explores language-guided visual feature extraction, where language priors directly modulate the visual backbone.
For instance, VG-LAW~\cite{su2023language} generates language-adaptive weights for the visual encoder to enable expression-relevant feature learning without heavy cross-modal fusion modules.
Nevertheless, most existing multi-task grounding methods are designed for natural-scene settings and do not account for the unique characteristics of remote sensing imagery, \emph{i.e.}, extreme scale variations, dense object layouts, and complex backgrounds.
To address this limitation, we propose ProVG, a remote sensing-oriented multi-task grounding framework that explicitly decouples heterogeneous linguistic cues into global context, spatial relations and object attributes and progressively integrates them to achieve accurate visual grounding.

\begin{table}[!t]
\centering
\caption{Comparison of different cross-modal modulators and cue-injection schemes on the RRSIS-D test set.
\textit{S}, \textit{L}, and \textit{V} denote \textit{\textcolor{mygreen}{Survey}}, \textit{\textcolor{myblue}{Locate}}, \textit{\textcolor{myorange}{Verify}} Attention, respectively.}
\resizebox{\linewidth}{!}{
\begin{tabular}{ll|ccc|ccc}
    \specialrule{.1em}{.1em}{.1em} 
    \multirow{2}{*}{Methods}                &       & \multicolumn{3}{c|}{RSREC} & \multicolumn{3}{c}{RSRES}\\
    \cmidrule(r){3-8}       
                                            &       & Pr@0.5 & oIoU & mIoU & Pr@0.5 & oIoU  & mIoU\\
    \specialrule{.1em}{.1em}{.1em} 
    Fig.~\ref{fig:2}\textcolor{mmpurple}{(a)}  &     & 72.66 & 73.45 & 63.92 & 74.06 & 74.56 & 64.01 \\
    Fig.~\ref{fig:2}\textcolor{mmpurple}{(b)}  &     & 71.07 & 71.89 & 63.77 & 71.75 & 72.23 & 62.00 \\
    Fig.~\ref{fig:2}\textcolor{mmpurple}{(c)}  &     & 72.54 & 72.12 & 64.31 & 73.47 & 73.89 & 63.75 \\
    \hline
    \multirow{4}{*}{Fig.~\ref{fig:2}\textcolor{mmpurple}{(d)}} & \textit{\textcolor{mygreen}{S}-\textcolor{myblue}{L}-\textcolor{myorange}{V}}   & \textbf{76.21} & \textbf{78.28} & \textbf{66.17} & \textbf{76.10} & \textbf{77.62} & \textbf{65.44} \\
                                                              & \textit{\textcolor{mygreen}{S}-\textcolor{myorange}{V}-\textcolor{myblue}{L}}   & 74.93 & 75.04 & 64.93 & 75.01 & 75.26 & 64.18 \\
                                                              & \textit{\textcolor{myblue}{L}-\textcolor{myorange}{V}-\textcolor{mygreen}{S}}   & 75.09 & 76.33 & 65.02 & 75.32 & 76.08 & 64.41 \\
                                                              & \textit{\textcolor{myorange}{V}-\textcolor{myblue}{L}-\textcolor{mygreen}{S}}   & 75.15 & 77.10 & 65.84 & 75.07 & 76.35 & 64.70 \\
    \specialrule{.1em}{.1em}{.1em} 
\end{tabular}}
\label{tab:modulator}
\end{table}

\section{Analyzing Cross-modal Modulator Designs for RSVG}
\label{sec3}
Prior work~\cite{li2024language,liu2024rotated} has shown that incorporating language guidance during visual feature extraction can significantly improve grounding performance, particularly in complex remote sensing scenes.
Despite encouraging progress, existing designs remain largely heuristic and lack a systematic understanding of how linguistic cues should be represented and injected.
This motivates a systematic investigation of cross-modal modulator designs for RSVG.

To enable this analysis, we formulate diverse cross-modal modulators under a unified attention-based view.
Specifically, visual features are treated as queries, while linguistic features serve as keys and values to modulate visual representations.
Under this formulation, modulation strategies can be characterized along two dimensions:
(i) how referring expressions are represented, and 
(ii) how heterogeneous linguistic cues are organized and injected into the visual backbone.
Based on this, we instantiate four representative modulators, as shown in Fig.~\ref{fig:2}:
\begin{itemize}[leftmargin=*]
  \item \textbf{Global context guidance.}
  A single sentence embedding is used to modulate visual features at each backbone stage, without explicitly distinguishing between different types of cues within the referring expression.
  \item \textbf{Parallel injection of decoupled spatial and attribute cues.}
  The referring expression is decomposed into spatial and attribute cues, which are encoded independently and injected in parallel to guide visual feature extraction.  
  \item \textbf{Sequential injection of decoupled spatial and attribute cues.}
  Spatial and attribute cues are injected sequentially, where spatial guidance first narrows down candidate regions, followed by attribute-based refinement.
  \item \textbf{Progressive cross-modal modulator.}
  Global context is introduced as an initial priming signal to establish global semantic awareness, followed by the sequential injection of spatial and attribute cues.
\end{itemize}
We empirically evaluate these strategies on the RRSIS-D test set, with results summarized in Tab.~\ref{tab:modulator}. 
Specifically, global context guidance (Fig.~\ref{fig:2}\textcolor{mmpurple}{(a)}) achieves 72.66 Pr@0.5 on RSREC and 64.01 mIoU on RSRES.
When decoupled spatial and attribute cues are injected in parallel (Fig.~\ref{fig:2}\textcolor{mmpurple}{(b)}), performance degrades significantly, with RSREC Pr@0.5 dropping to 71.07 and RSRES mIoU to 62.00.
Sequential injection (Fig.~\ref{fig:2}\textcolor{mmpurple}{(c)}) recovers performance to a level comparable with the single-sentence baseline, achieving 72.54 Pr@0.5 and 63.75 mIoU.
When global context is further incorporated and cues are progressively integrated (Fig.~\ref{fig:2}\textcolor{mmpurple}{(d)}), performance is consistently improved across all metrics. 
These results suggest that effective grounding benefits from both structured cue organization and progressive integration of heterogeneous linguistic cues.

\begin{figure}[!t]
    \centering
    \includegraphics[width=0.99\linewidth]{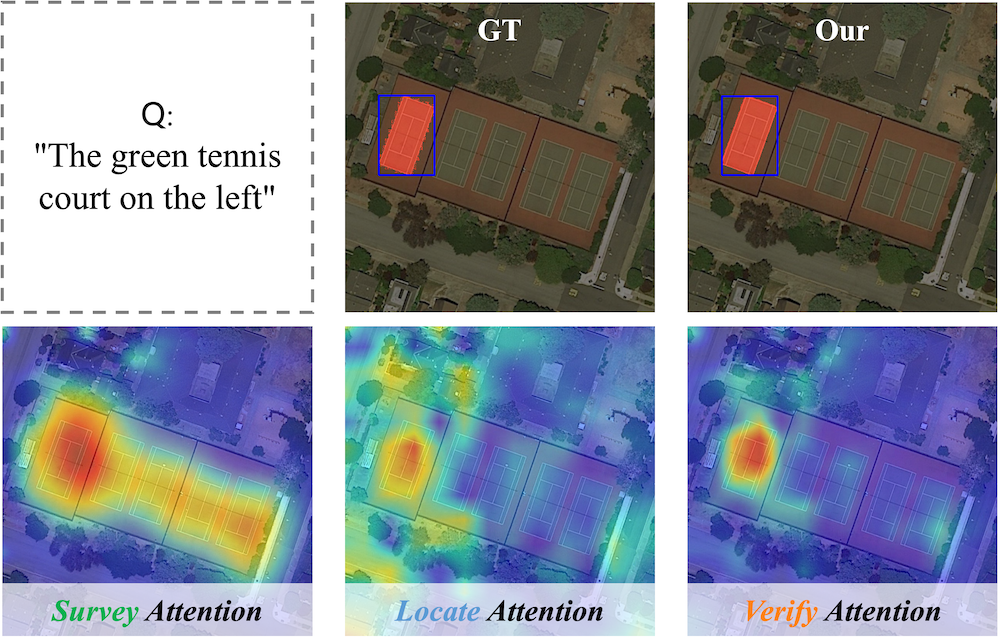}
    \caption{Visualization of the attention maps of the proposed progressive cross-modal modulator.}
    \label{fig:LPVatten}
\end{figure}

We further investigate the impact of cue-injection order within the progressive modulator by enumerating different permutations of
\emph{\textcolor{mygreen}{Survey}} \emph{(\textcolor{mygreen}{S})},
\emph{\textcolor{myblue}{Locate}} \emph{(\textcolor{myblue}{L})}, and
\emph{\textcolor{myorange}{Verify}} \emph{(\textcolor{myorange}{V})}, as reported in Tab.~\ref{tab:modulator}.
The \textit{\textcolor{mygreen}{S}-\textcolor{myblue}{L}-\textcolor{myorange}{V}} ordering achieves the best overall performance, reaching \textbf{76.21/66.17} Pr@0.5/mIoU on RSREC and \textbf{76.10/65.44} Pr@0.5/mIoU on RSRES.
In contrast, alternative orders consistently yield inferior results, with RSREC Pr@0.5 ranging from 74.93 to 75.15 and RSRES mIoU ranging from 64.18 to 64.70.
To further illustrate the grounding process, we visualize the attention maps of the proposed modulator in Fig.~\ref{fig:LPVatten}.
The visualization reveals a progressive grounding process: a global understanding is first established, followed by spatial narrowing of candidate regions, and finally attribute-based verification of the target, aligning with the staged perception process observed in human object grounding.

\begin{figure*}[!t]
    \centering
    \includegraphics[width=\linewidth]{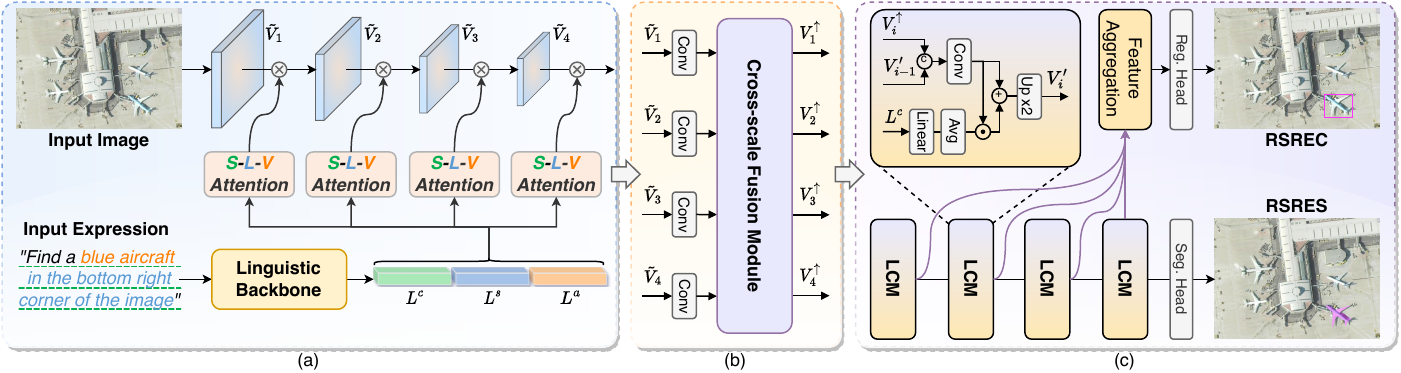}
    \caption{Overall framework of the proposed ProVG.
    It consists of three components: 
    (a) a visual--text feature extractor with progressive cross-modal modulator, 
    (b) a cross-scale fusion module, and 
    (c) a language-guided calibration decoder with a unified multi-task prediction head.}
    \label{fig:framework}
\end{figure*}

\section{Proposed Framework}
\label{sec4}
\subsection{Overview}
The overall architecture of ProVG is illustrated in Fig.~\ref{fig:framework}.
Given an expression $E$, we first decompose it into spatial cues and attribute cues using NLTK~\cite{loper2002nltk}, and encode them with BERT~\cite{devlin2018bert} to obtain spatial linguistic features $L^s \in \mathbb{R}^{N_s \times D}$ and attribute linguistic features $L^a \in \mathbb{R}^{N_a \times D}$.
In parallel, the full expression is encoded to produce a global linguistic representation $L^c \in \mathbb{R}^{N_c \times D}$.
For the image $I$, we adopt a Swin Transformer~\cite{liu2021swin} to extract multi-scale visual features $\{V_i\}_{i=1}^{4}$, where $V_i \in \mathbb{R}^{H_i W_i \times C_i}$.
These features are progressively modulated by $L^c$, $L^s$, and $L^a$ via the proposed Progressive Cross-modal Modulator (PCM), yielding language-modulated features $\{\tilde{V}_i\}_{i=1}^{4}$.
The modulated features are then processed by a Cross-scale Fusion Module (CFM) to enhance cross-scale interaction.
Subsequently, a Language-guided Calibration Decoder (LCM) further refines the fused representations and performs task-specific prediction for both RSREC and RSRES.
The entire network is trained with a joint multi-task objective.

\subsection{Progressive Cross-modal Modulator}
\label{sec:pcm}
Motivated by the observations in Sec.~\ref{sec3}, we propose a simple yet effective Progressive Cross-modal Modulator (PCM) that enables fine-grained vision-language interaction at each feature scale.
Concretely, at each backbone stage, PCM generates three adaptive modulation weights for visual features ${V}_i$, including context weights $W_i^{c}$, spatial weights $W_i^{s}$, and attribute weights $W_i^{a}$.
These weights are applied sequentially to progressively constrain the feature space, forming a coarse-to-fine reasoning pipeline.

\subsubsection{\textbf{Survey Attention.}}
The survey attention module is designed to establish global semantic awareness by leveraging sentence-level contextual information.
Given the visual features $V_i \in \mathbb{R}^{H_i W_i \times C_i}$ and global linguistic features $L^c \in \mathbb{R}^{N_c \times D}$, the module applies two $1 \times 1$ convolutions to project them to the same channel dimension $C_i$, yielding $\tilde{V}_i \in \mathbb{R}^{H_iW_i \times C_i}$ and $\tilde{{L}}^c \in \mathbb{R}^{N_c \times C_i}$. 
Cross-modal attention is then performed using $\tilde{V}_i$ as queries and $\tilde{L}^c$ as keys and values:
\begin{equation}
  A_i^{c} = \mathrm{Attention}(W_q^{c} \tilde{V}_i, W_k^{c} \tilde{L}_c, W_v^{c} \tilde{L}^c),
\end{equation}
where $W_q^{c}$, $W_k^{c}$, and $W_v^{c}$ denote learnable projection matrices.
The output $A_i^{c} \in \mathbb{R}^{H_i W_i \times C_i}$ is transformed into a modulation weight $W_i^{c} \in \mathbb{R}^{H_i W_i \times C_i}$ via a Sigmoid function and applied to the original visual features:
\begin{equation}
V_i^{c} = \mathrm{Proj}(V_i \odot W_i^{c}),\quad W_i^{c} = \sigma(A_i^{c}),
\end{equation}
where $\odot$ denotes element-wise multiplication and $\sigma (\cdot)$ is the Sigmoid function.
The resulting result is passed through a $1\times 1$ convolution, denoted as $\mathrm{Proj}(\cdot)$, to obtain the context-aware visual features ${V}_{i}^{c} \in \mathbb{R}^{H_iW_i \times C_i}$ that encode global semantic intent and provide a reliable prior for subsequent localization.

\subsubsection{\textbf{Locate Attention.}}
After obtaining $V_i^{c}$, the locate attention module leverages spatial cues to further narrow down candidate regions.
Concretely, $V_i^{c}$ is transformed via a $1 \times 1$ convolution into a spatial representation $\tilde{V}_i^{s} \in \mathbb{R}^{H_i W_i \times 1}$ that encodes spatial saliency.
The spatial linguistic features $L^s$ are projected to $\tilde{L}^s \in \mathbb{R}^{N_s \times 1}$ to align with this representation. 
The module then performs attention using $\tilde{V}_i^{s}$ as queries and $\tilde{L}^s$ as keys and values:
\begin{equation}
A_i^{s} = \mathrm{Attention}(W_q^{s} \tilde{V}_i^{s}, W_k^{s} \tilde{L}^s, W_v^{s} \tilde{L}^s),
\end{equation}
where $A_i^{s} \in \mathbb{R}^{H_i W_i \times 1}$ reflects the relevance between each visual location and the described spatial relations.
The attention map is converted into a spatial modulation weight $W_i^{s} = \sigma(A_i^{s})$ and applied to $V_i^{c}$ as:
\begin{equation}
V_i^{s} = \mathrm{Proj}(V_i^{c} \odot W_i^{s}),
\end{equation}
where $V_i^{s}$ denotes spatially-aware features that emphasize regions consistent with the described spatial relations, thereby narrowing the search space before attribute verification.

\subsubsection{\textbf{Verify Attention.}}
Despite spatial filtering, objects within candidate regions may still exhibit similar spatial patterns, leaving residual ambiguity.
To resolve this, we perform attribute-level verification to generate a channel-wise modulation weight.
Specifically, given the features $V_i^{s} \in \mathbb{R}^{H_iW_i \times C_i}$, we first apply mean pooling to obtain a compact channel descriptor $\tilde{V}_i^{a} = \mathrm{MeanPool}(V_i^{s}) \in \mathbb{R}^{1 \times C_i}$, which summarizes the attribute- and category-related responses of the localized candidates.
To enable independent interaction between each channel and the attribute words, we transpose $\tilde{V}_i^{a}$ to treat each channel as a query token:
\begin{equation}
\bar{V}_i^{a} = (\tilde{V}_i^{a})^{\mathsf{T}} \in \mathbb{R}^{C_i \times 1}.
\end{equation}
The attribute linguistic features $L^a$ are projected to align with the query dimension, yielding $\tilde{L}^a \in \mathbb{R}^{N_a \times 1}$.
Similar to the previous modules, we take $\bar{V}_i^{a}$ as queries and $\tilde{L}^a$ as keys and values to perform cross-modal attention:
\begin{equation}
A_i^{a} = \mathrm{Attention}(W_q^{a} \bar{V}_i^{a},\; W_k^{a} \tilde{L}^a,\; W_v^{a} \tilde{L}^a),
\end{equation}
where the output $A_i^{a} \in \mathbb{R}^{C_i \times 1}$ captures channel-specific aggregated attribute information.
Finally, we obtain the channel-wise modulation weight via sigmoid and transpose it back to produce the final modulated features:
\begin{equation}
\tilde{V}_i = V_i^{s} \odot \sigma\bigl((A_i^{a})^{\mathsf{T}}\bigr), \quad \tilde{V}_i \in \mathbb{R}^{H_iW_i \times C_i}.
\end{equation}

\subsection{Cross-scale Fusion Module}
\label{sec:cfm}
While PCM effectively captures multi-scale feature extraction under heterogeneous linguistic cue guidance, we additionally design a Cross-scale Fusion Module (CFM) to facilitate the bidirectional information exchange between the coarse and fine stages, particularly for handling large scale variations in remote sensing images.
Specifically, 
CFM takes the multi-scale features from PCM, \textit{i.e.}, the previously mentioned $\tilde{V}_i,i\in\{1,2,3,4\}$ as input and performs bidirectional interaction.
\subsubsection{\textbf{Top-down Interaction.}}
We first propagate high-level semantic information from coarse to fine scales, enabling fine-resolution features to inherit strong contextual priors.
For $\tilde{V}_i$ with $i \in \{3,2,1\}$, we aggregate information from adjacent higher-level features:
\begin{equation}
{V}_i^{\downarrow} = \mathrm{FusionBlock} ( \tilde{{V}}_i + \mathrm{Up}(\tilde{{V}}_{i+1}) ),
\end{equation}
where $\mathrm{Up}(\cdot)$ denotes upsampling, and $\mathrm{FusionBlock}(\cdot)$ is a lightweight re-parameterizable convolution block~\cite{zhao2024detrs} for cross-scale feature fusion.
This operation enhances fine-scale features by injecting high-level semantic context, improving their ability to capture semantically relevant regions.

\subsubsection{\textbf{Bottom-up Interaction.}}
To complement the semantic propagation, we further introduce a bottom-up interaction to propagate fine-grained spatial details back to coarse scales.
The bottom-up interaction is formulated as:
\begin{equation}
{V}_i^{\uparrow} = \mathrm{FusionBlock} ( {V}_i^{\downarrow} + \mathrm{Down}({V}_{i-1}^{\downarrow}) ),
\quad i = \{2,3,4\},
\end{equation}
where $\mathrm{Down}(\cdot)$ denotes spatial reduction via a $3 \times 3$ convolution with stride 2.
This process enables detailed localization cues from fine-resolution features to refine higher-level representations, enhancing their sensitivity to precise object structures.
The output of CFM is then passed to the decoder for final box and mask prediction.

\begin{table*}[!t]
\centering
\caption{Comparisons with SOTA methods on \textbf{RRSIS-D} test set.
$\dagger$ indicates results from original papers.
\textcolor{mygray}{Gray} indicates results obtained using a vision-language model.
\tbredbf{Best} and \tbbluebf{Second Best} performances are highlighted.}
\resizebox{\linewidth}{!}{
\begin{tabular}{l|ccccccc|ccccccc}
    \specialrule{.1em}{.1em}{.1em} 
    \multirow{2}{*}{Methods}     & \multicolumn{7}{c|}{RSREC}                                 & \multicolumn{7}{c}{RSRES}\\
    \cmidrule(r){2-15}       
                                 & Pr@0.5 & Pr@0.6 & Pr@0.7 & Pr@0.8 & Pr@0.9 & oIoU  & mIoU  & Pr@0.5 & Pr@0.6 & Pr@0.7 & Pr@0.8 & Pr@0.9 & oIoU  & mIoU\\
    \specialrule{.1em}{.1em}{.1em}  
    \multicolumn{2}{l}{\textbf{\textsl{VLM methods:}}}\\
    $\dagger$GeoGround~\cite{zhou2024geoground}          & \g{-}     & \g{-}     & \g{-}     & \g{-}     & \g{-}     & \g{-}     & \g{-}     & \g{67.50} & \g{-}     & \g{-}     & \g{-}     & \g{-}     & \g{-}     & \g{60.50} \\
    $\dagger$SegEarth-R1~\cite{li2025segearth_r1}        & \g{-}     & \g{-}     & \g{-}     & \g{-}     & \g{-}     & \g{-}     & \g{-}     & \g{76.96} & \g{71.88} & \g{61.62} & \g{47.17} & \g{28.18} & \g{78.01} & \g{66.40} \\
    $\dagger$Tex4Seg++~\cite{lan2025text4seg++}          & \g{-}     & \g{-}     & \g{-}     & \g{-}     & \g{-}     & \g{-}     & \g{-}     & \g{73.20} & \g{-}     & \g{-}     & \g{-}     & \g{-}     & \g{74.20} & \g{62.80} \\
    $\dagger$RSVG-ZeroOV~\cite{li2026rsvg}               & \g{31.39} & \g{24.15}     & \g{17.63} & \g{11.43}     & \g{4.33}     & \g{31.28} & \g{34.49} & \g{27.39} & \g{20.56}     & \g{13.38} & \g{6.95}     & \g{2.26}     & \g{22.83} & \g{28.35} \\
    \hline
    \multicolumn{2}{l}{\textbf{\textsl{REC methods:}}}\\
    QRNet~\cite{ye2022shifting}                          & 72.63 & 63.87 & 58.61 & 45.35 & 21.72 & 73.48 & 61.34 & -     & -     & -     & -     & -     & -     & -     \\
    VLTVG~\cite{yang2022improving}                       & 72.86 & 65.93 & 60.56 & 48.14 & 26.21 & 72.91 & 63.51 & -     & -     & -     & -     & -     & -     & -     \\
    MGVLF~\cite{zhan2023rsvg}                            & 71.05 & 66.17 & 59.89 & 48.36 & 26.75 & 71.28 & 61.44 & -     & -     & -     & -     & -     & -     & -     \\
    TransCP~\cite{tang2023context}                       & 30.56 & 21.36 & 13.49 &  4.74 &  0.63 & -     & 29.12 & -     & -     & -     & -     & -     & -     & -     \\
    LPVA~\cite{li2024language}                           & 74.32 & 68.45 & 61.12 & 50.74 & 28.62 & 74.29 & 64.82 & -     & -     & -     & -     & -     & -     & -     \\
    LQVG~\cite{lan2024language}                          & \tbbluebf{75.63} & \tbbluebf{70.67} & \tbbluebf{63.77} & \tbbluebf{51.91} & \tbbluebf{29.93} & \tbbluebf{75.35} & \tbbluebf{66.07} & -     & -     & -     & -     & -     & -     & -     \\
    \hline
    \multicolumn{2}{l}{\textbf{\textsl{RES methods:}}}\\
    LAVT-RIS~\cite{yang2022lavt}                         & 68.03 & 63.52 & 56.46 & 44.78 & 28.51 & 72.67 & 64.98 & 69.48 & 63.10 & 52.97 & 41.07 & \tbbluebf{24.26} & 76.48 & 61.12 \\
    LAVT-RS~\cite{10694805}                              & 68.12 & 63.47 & 56.69 & 45.11 & 28.54 & 72.86 & 65.04 & 69.82 & 63.25 & 53.02 & 41.35 & 24.17 & 76.39 & 61.44 \\
    LGCE~\cite{yuan2023rrsis}                            & 67.97 & 62.08 & 54.55 & 44.73 & 27.98 & 72.24 & 64.47 & 69.72 & 62.80 & 52.71 & 40.42 & 23.81 & 76.33 & 60.98 \\
    RMISN~\cite{liu2024rotated}                          & 73.02 & 66.16 & 58.06 & 46.34 & 28.76 & 73.94 & 65.97 & 74.03 & 66.93 & 55.44 & 41.65 & 23.07 & 76.55 & 63.38 \\
    RefSegformer~\cite{wu2024toward}                     & 66.58 & 61.15 & 54.09 & 43.50 & 26.53 & 72.32 & 59.81 & 69.25 & 63.48 & 53.06 & 40.31 & 24.03 & \tbbluebf{77.44} & 59.67 \\
    FIANet~\cite{lei2024exploring}                       & 73.02 & 66.82 & 58.92 & 46.88 & 28.61 & 73.88 & 65.01 & 74.35 & \tbbluebf{66.96} & \tbbluebf{56.25} & \tbbluebf{42.66} & 23.93 & 76.81 & 64.01 \\ 
    $\dagger$CroBIM~\cite{dong2024cross}                 & -     & -     & -     & -     & -     & -     & -     & \tbbluebf{75.00} & 66.32 & 54.31 & 41.09 & 21.78 & 76.37 & \tbbluebf{64.24} \\ 
    \hline
    \multicolumn{2}{l}{\textbf{\textsl{Multi-task methods:}}}\\  
    RefTR~\cite{li2021referring}                         & 30.16 & 24.03 & 15.93 &  7.50 &  0.77 & 45.20 & 28.10 & -     & -     & -     & -     & -     & -     & -     \\
    EEVG~\cite{chen2024efficient}                        & 61.75 & 55.34 & 45.49 & 30.90 &  8.96 & 74.18 & 54.25 & 70.02 & 61.55 & 49.89 & 36.56 & 17.23 & 74.53 & 61.11  \\
    \textbf{ProVG (Ours)}                                & \tbredbf{76.21} & \tbredbf{70.84} & \tbredbf{63.78} & \tbredbf{52.14} & \tbredbf{30.74} & \tbredbf{78.28} & \tbredbf{66.17} & \tbredbf{76.10} & \tbredbf{70.30} & \tbredbf{58.52} & \tbredbf{44.47} & \tbredbf{25.91} & \tbredbf{77.62} & \tbredbf{65.44} \\
    \specialrule{.1em}{.1em}{.1em}  
\end{tabular}}
\label{tab:RRSIS-D}
\end{table*}

\subsection{Language-guided Calibration Decoder}
\label{sec:lcd}
To better align visual features with downstream prediction objectives, we introduce a Language-guided Calibration Decoder (LCD) that performs stage-wise refinement of decoding representations under global linguistic guidance.
Specifically, LCD takes the cross-scale fused features $\{V_i^{\uparrow}\}_{i=1}^{4}$ as input and applies a series of language-guided calibration modules (LCMs) to enhance the decoding features.
We further design a task-specific prediction head to accommodate different downstream objectives.
\subsubsection{\textbf{Language-guided Calibration Module.}}
At decoding stage $i$, we integrate the current-scale feature $V_i^{\uparrow}$ with the previous decoded feature ${V}'_{i-1}$ to refine the decoding representation.
The fused feature $X_i$ is obtained by concatenation followed by a $1 \times 1$ convolution:
\begin{equation}
{X}_i =
\begin{cases}
\mathrm{Conv}({V}_4^{\uparrow}), & i=4\\
\mathrm{Conv}(\mathrm{Cat}({V}_i^{\uparrow},\, {V}'_{i-1})), & i\in\{3,2,1\}
\end{cases}.
\end{equation}
We generate a language-conditioned calibration gate from the global linguistic representation $L^c$ and use it to reweight the fused feature, followed by upsampling to obtain the decoded output. 
The process is formulated as:
\begin{equation}
{V}'_i = \mathrm{Up}(\tilde{{X}}_i), \quad
\tilde{{X}}_i = {X}_i + {X}_i \odot 
\sigma\!\left(\mathrm{AvgPool}(\mathrm{Linear}({L}^c))\right).
\end{equation}
\subsubsection{\textbf{Task-specific Prediction.}}
Considering that RSREC and RSRES impose different representation requirements, we design a task-specific prediction head.

For RSRES, the final decoded feature ${V}'_4$ is directly projected into class score maps via a $1\times1$ convolution for mask prediction.

For RSREC, we aggregate multi-scale decoding features to exploit complementary information across stages.
Specifically, the features from all decoding stages are first projected to a unified channel dimension and stacked as $F = [V'_1, V'_2, V'_3, V'_4] \in \mathbb{R}^{B \times 4 \times C}$.
A stage descriptor is obtained by averaging along the channel dimension, followed by a linear layer and softmax to produce adaptive weights:
\begin{equation}
{w} = \mathrm{Softmax}(\mathrm{Linear}(\mathrm{Mean}({F}))),
\quad {w}\in\mathbb{R}^{B\times4}.
\end{equation}
The aggregated representation is computed as:
\begin{equation}
{F}_{rec}={\textstyle \sum_{i=1}^{4}}({w}_i \odot {F}_i),
\end{equation}
which is fed into a three-layer MLP to predict the box coordinates.

\subsection{Training Objective}
\label{sec:loss}
Following prior works~\cite{zhan2023rsvg,liu2024rotated}, we employ the smooth L1 loss and the generalized intersection-over-union (GIoU) loss on the 4-D bounding box coordinates for RSREC, denoted as $\mathcal{L}_{\text{box}}$. 
For RSRES, we adopt the standard cross-entropy loss and Dice loss, denoted as $\mathcal{L}_{\text{mask}}$.
To enforce geometric consistency between the predicted box and mask, we introduce an additional constraint loss $\mathcal{L}_{\text{cons}}$. 
Concretely, we derive a bounding box $\mathbf{b}_m$ from the predicted mask by computing its minimal enclosing rectangle. 
This box is aligned with the predicted bounding box $\mathbf{b}_p$ using both smooth L1 and GIoU losses as:
\begin{equation}
\mathcal{L}_{\text{cons}} = 10\,\mathcal{L}_{\text{smooth-L1}}(\mathbf{b}_p,\mathbf{b}_m) + \mathcal{L}_{\text{GIoU}}(\mathbf{b}_p,\mathbf{b}_m).
\end{equation}
The overall training objective is defined as:
\begin{equation}
\mathcal{L} = \lambda_1 \mathcal{L}_{\text{box}} + \lambda_2 \mathcal{L}_{\text{mask}} + \lambda_3 \mathcal{L}_{\text{cons}},
\end{equation}
where $\lambda_1=1$, $\lambda_2=0.5$, and $\lambda_3=0.1$ are hyperparameters that balance the contributions of different loss terms.

\begin{table*}[!t]
\centering
\caption{Comparisons with SOTA methods on \textbf{RISBench} test set. 
$\dagger$ indicates results from original papers.
\textcolor{mygray}{Gray} indicates results obtained using a vision-language model.
\tbredbf{Best} and \tbbluebf{Second Best} performances are highlighted.}
\resizebox{\linewidth}{!}{
\begin{tabular}{l|ccccccc|ccccccc}
    \specialrule{.1em}{.1em}{.1em}  
    \multirow{2}{*}{Methods}                     & \multicolumn{7}{c|}{RSREC}                                 & \multicolumn{7}{c}{RSRES}\\
    \cmidrule(r){2-15}       
                                                 & Pr@0.5 & Pr@0.6 & Pr@0.7 & Pr@0.8 & Pr@0.9 & oIoU  & mIoU  & Pr@0.5 & Pr@0.6 & Pr@0.7 & Pr@0.8 & Pr@0.9 & oIoU  & mIoU\\
    \specialrule{.1em}{.1em}{.1em}  
    \textbf{\textsl{VLM methods:}}\\
    $\dagger$RSVG-ZeroOV~\cite{li2026rsvg}        & \g{38.90} & \g{31.93}     & \g{24.93} & \g{18.66}     & \g{8.78}     & \g{34.30} & \g{38.87} & \g{31.03} & \g{24.43}     & \g{18.61} & \g{12.48}     & \g{5.49}     & \g{26.35} & \g{31.84} \\
    \hline
    \multicolumn{2}{l}{\textbf{\textsl{REC methods:}}}\\
    QRNet~\cite{ye2022shifting}                  & 69.35 & 66.51 & 62.08 & 52.43 & 35.29 & 62.03 & 59.84 & -     & -     & -     & -     & -     & -     & -     \\
    VLTVG~\cite{yang2022improving}               & 68.16 & 66.17 & 61.54 & 52.40 & 35.00 & 61.60 & 59.39 & -     & -     & -     & -     & -     & -     & -     \\
    MGVLF~\cite{zhan2023rsvg}                    & 67.84 & 65.92 & 61.03 & 52.11 & 34.67 & 61.38 & 59.21 & -     & -     & -     & -     & -     & -     & -     \\
    TransCP~\cite{tang2023context}               & 29.87 & 22.14 & 14.92 &  6.08 &  1.27 & 42.29 & 33.93 & -     & -     & -     & -     & -     & -     & -     \\
    LPVA~\cite{li2024language}                   & 72.91 & 68.47 & 64.88 & 55.02 & 36.41 & 64.52 & 65.97 & -     & -     & -     & -     & -     & -     & -     \\
    LQVG~\cite{lan2024language}                  & \tbbluebf{75.26} & \tbbluebf{71.31} & \tbbluebf{66.17} & \tbbluebf{57.33} & \tbredbf{38.79} & 67.09 & \tbbluebf{67.93}	& -     & -     & -     & -     & -     & -     & -     \\
    \hline
    \multicolumn{2}{l}{\textbf{\textsl{RES methods:}}}\\
    LAVT-RIS~\cite{yang2022lavt}                 & 69.96 & 65.55 & 59.60 & 50.45 & 32.71 & 72.88 & 65.93	& 69.52 & 63.63 & 56.10 & 44.95 & 25.21 & 74.15 & 61.93 \\
    LAVT-RS~\cite{10694805}                      & 69.48 & 66.31 & 60.02 & 50.63 & 32.94 & \tbbluebf{73.06} & 65.98 & 69.81 & 64.02 & 55.96 & 45.03 & 25.68 & 74.06 & 62.11 \\
    LGCE~\cite{yuan2023rrsis}                    & 68.03 & 64.97 & 57.28 & 51.64 & 30.16 & 72.91 & 64.80& 69.64 & 64.07 & 56.26 & 44.92 & 25.74 & 73.87 & 62.13 \\
    RMISN~\cite{liu2024rotated}                  & 72.97 & 68.64 & 62.72 & 52.39 & 32.65 & 70.11 & 66.90 & \tbbluebf{76.40} & \tbbluebf{71.07} & 53.51 & 52.21 & 30.48 & 74.42 & \tbbluebf{67.88}\\
    RefSegformer~\cite{wu2024toward}             & 65.73 & 61.16 & 55.14 & 45.89 & 29.00 & 68.89 & 60.51 & 68.16 & 62.83 & 55.16 & 43.60 & 24.32 & 73.89 & 60.83\\
    FIANet~\cite{lei2024exploring}               & 73.30 & 68.77 & 63.12 & 52.99 & 33.72 & 71.32 & 66.84 & 75.73 & 70.65 & \tbbluebf{63.85} & \tbbluebf{53.27} & \tbbluebf{32.13} & \tbbluebf{74.49} & 67.74 \\ 
    $\dagger$CroBIM~\cite{dong2024cross}         & -   & -     & -      & -     & -     & -    & -   & 75.75 & 70.34 & 63.12 & 51.12 & 28.45 & 73.61 & 67.32 \\ 
    \hline
    \multicolumn{2}{l}{\textbf{\textsl{Multi-task methods:}}}\\  
    RefTR~\cite{li2021referring}                 & 34.38 & 29.39 & 21.04 &  11.70 &  2.15 & 45.80 & 32.16 & -     & -     & -     & -     & -     & -     & -     \\
    EEVG~\cite{chen2024efficient}                & 68.48 & 60.11 & 48.99 & 33.16 & 12.52 & 70.70 & 58.97 & 70.42 & 61.62 & 49.52 & 33.76 & 16.44 & 72.70 & 60.97 \\
    \textbf{ProVG (Ours)}                        & \tbredbf{78.06} & \tbredbf{74.29} & \tbredbf{68.69} & \tbredbf{58.57} & \tbbluebf{38.23} & \tbredbf{75.69} & \tbredbf{70.82} & \tbredbf{79.02} & \tbredbf{74.68} & \tbredbf{68.42} & \tbredbf{57.73} & \tbredbf{37.01} & \tbredbf{75.86} & \tbredbf{70.63}  \\
    \specialrule{.1em}{.1em}{.1em}  
\end{tabular}}
\label{tab:RISBench}
\end{table*}

\section{Experiments}
\subsection{Experimental Settings}
\textbf{Datasets.}
We evaluate our method on two benchmarks tailored for RSVG: RRSIS-D \cite{liu2024rotated} and RISBench \cite{dong2024cross}.
Each sample comprises an RGB image, a referring expression, and a segmentation mask. 
Both datasets are built upon RSREC benchmarks and inherit their bounding box annotations, enabling comprehensive evaluation of both RSREC and RSRES tasks.

\noindent \textbf{Baselines.}
We compare our method with a wide range of state-of-the-art approaches, including 4 VLM methods, 6 REC methods, 7 RES methods, and 2 multi-task grounding methods.

\noindent \textbf{Implementation Details.}
For fair comparison, we follow the training settings reported in the original papers for all baselines.
Our model is trained for 70 epochs using the AdamW optimizer with a weight decay of 0.01 and an initial learning rate of $3\times10^{-5}$.
The learning rate is decayed by a factor of 0.1 after the 50th and 60th epochs, and training is terminated at epoch 70.
Following prior works~\cite{li2024language, liu2024rotated}, we report mIoU, oIoU, and Precision@X with $X \in \{0.5, 0.6, 0.7, 0.8, 0.9\}$.
Note that for RSRES, bounding boxes are derived from the predicted masks by computing their minimal enclosing rectangles.
All experiments are conducted on two NVIDIA GeForce RTX 4090 GPUs with 24 GB of VRAM.

\subsection{Main Results}
\noindent \textbf{Comparison Results on RRSIS-D.}
Tab.~\ref{tab:RRSIS-D} compares ProVG with existing REC, RES, and multi-task grounding methods on RRSIS-D. 
ProVG achieves the best performance on both RSREC and RSRES, consistently outperforming all prior approaches across all evaluation metrics.
For RSREC, ProVG improves the strongest REC baseline LQVG by \textbf{+2.93\%} in oIoU, indicating more accurate grounding of the referred objects. 
Compared with LPVA, which also employs cross-modal modulation during feature extraction, ProVG further gains \textbf{+1.89\%} in Pr@0.5 and \textbf{+1.35\%} in mIoU, demonstrating the effectiveness of progressively injecting decoupled context, spatial, and attribute cues over holistic feature conditioning.
For RSRES, ProVG also achieves the best performance, surpassing FIANet and CroBIM by \textbf{+1.43\%} and \textbf{+1.20\%} in mIoU, respectively, indicating more precise region prediction. 
Notably, ProVG outperforms GeoGround and Text4Seg++, which rely on large-scale vision-language models (\emph{e.g.}, 7B LLMs), while maintaining a significantly more lightweight architecture, highlighting the efficiency of the proposed framework.
ProVG also demonstrates clear advantages over previous multi-task grounding methods, achieving improvements of \textbf{+11.92\%} on RSREC and \textbf{+4.33\%} on RSRES in mIoU. 
These results validate the benefit of explicitly modeling heterogeneous semantic cues within a unified grounding framework.

\noindent \textbf{Comparison Results on RISBench.} 
We further evaluate ProVG on RISBench, a more challenging benchmark featuring longer referring expressions, richer spatial relations, and more cluttered scenes. 
As shown in Tab.~\ref{tab:RISBench}, ProVG consistently achieves the best results on both RSREC and RSRES, outperforming all competing methods on nearly all metrics.
In particular, ProVG improves mIoU by \textbf{+2.89\%} on RSREC and \textbf{+2.75\%} on RSRES over the strongest competitors, LQVG and RMSIN, respectively. 
Under the strict Pr@0.9 threshold, ProVG surpasses the second-best method FIANet by \textbf{4.88\%}. 
Compared with previous multi-task grounding approaches, ProVG also achieves substantial gains of \textbf{+9.58\%} and \textbf{+8.60\%} in Pr@0.5 on RSREC and RSRES.
These results suggest that the proposed decoupled linguistic modeling strategy remains robust under more complex referring expressions and challenging remote sensing scenarios.

\begin{figure*}[!t]
    \centering
    \includegraphics[width=\linewidth]{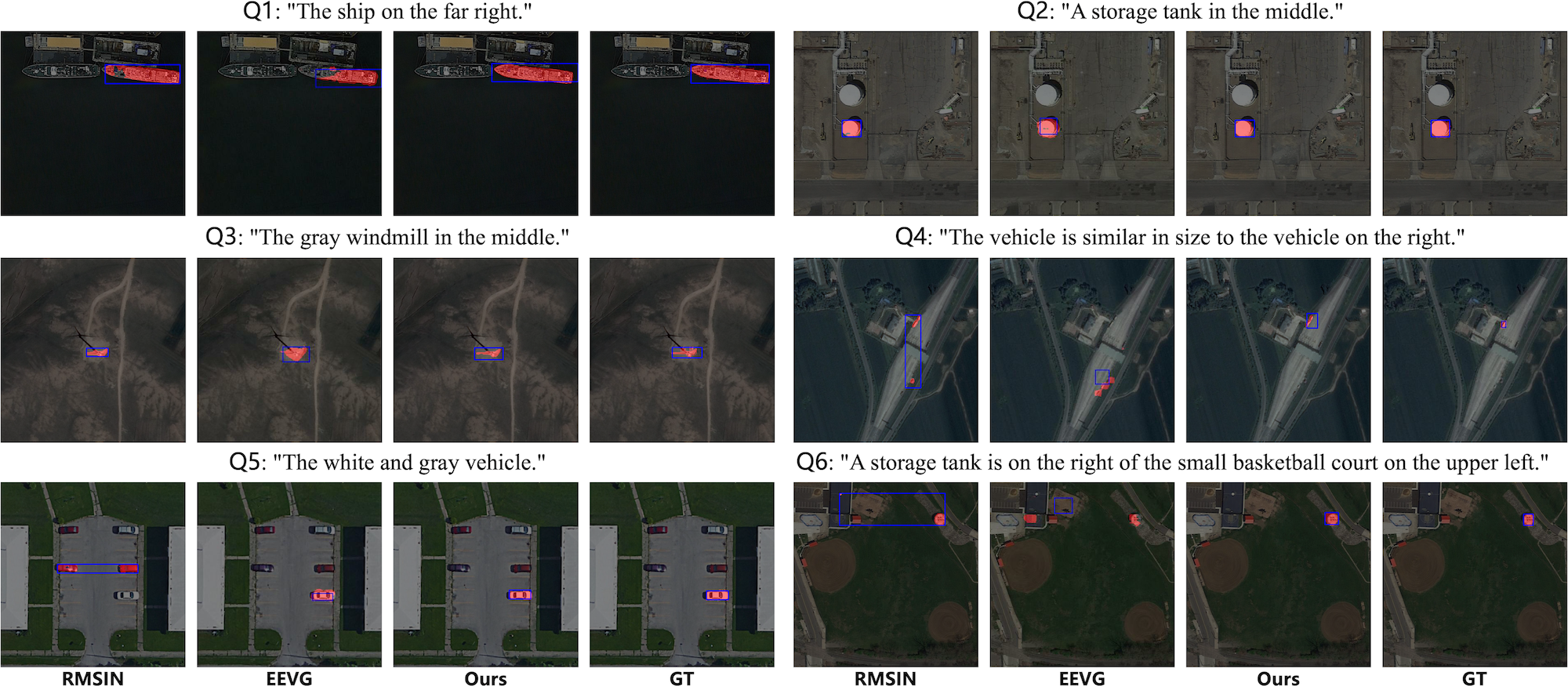}
    \caption{Qualitative comparisons between ProVG and previous SOTA methods.}
    \label{fig:5}
\end{figure*}

\noindent \textbf{Visual Comparisons.} 
Fig.~\ref{fig:5} compares ProVG with previous SOTA methods on both RSREC and RSRES tasks. 
In \textbf{Q1} and \textbf{Q2}, ProVG produces more complete segmentations that better cover the entire target, whereas RMSIN and EEVG often yield fragmented or incomplete masks. 
For small objects or fine structures (\textbf{Q3}), ProVG preserves sharper boundaries and avoids over-smoothed predictions, demonstrating improved spatial precision. 
More challenging relational expressions (\textbf{Q4-6}) further highlight the advantage of our design. 
Previous methods tend to produce inconsistent grounding results with mismatched box and mask, while ProVG maintains accurate alignment and delivers more reliable predictions.

\begin{table}[!t]
\centering
\caption{Effect of components in ProVG.}
\resizebox{\linewidth}{!}{
\begin{tabular}{ll|ccc|ccc}
    \toprule 
    \multicolumn{2}{c|}{}       & \multicolumn{3}{c|}{RSREC} & \multicolumn{3}{c}{RSRES}\\
    \cmidrule(r){3-8}       
    \multicolumn{2}{l|}{Exp.}           & Pr@0.5 & oIoU & mIoU & Pr@0.5 & oIoU & mIoU\\
    \midrule
    \multicolumn{2}{l|}{ProVG (Ours)}     & \textbf{76.21} & \textbf{78.28} & \textbf{66.17} & \textbf{76.10} & \textbf{77.62} & \textbf{65.44}   \\
    \multicolumn{2}{l|}{\textit{w/o} PCM} & 69.45 & 71.23 & 64.89 & 69.76 & 75.12 & 60.45 \\
    \multicolumn{2}{l|}{\textit{w/o} CFM} & 74.19 & 74.20 & 65.34 & 74.37 & 76.64 & 63.81 \\
    \multirow{2}{*}{LCD}      
    & \textit{w/o} LCM                    & 72.15 & 73.43 & 65.08 & 72.84 & 75.62 & 61.27 \\
    & \textit{w/o} FA                     & 75.46 & 76.94 & 65.25 & 75.86 & 77.05 & 65.21 \\
    \bottomrule 
\end{tabular}}
\label{tab:ablation1}
\end{table}

\subsection{Ablation Study}
\label{sec4c}
In this section, we report ablation study results on the RRSISD test set to investigate its effectiveness
.
\noindent \textbf{Effect of ProVG components.}
Tab.~\ref{tab:ablation1} evaluates the contribution of each component in ProVG. 
Removing PCM results in the largest performance drop on both RSREC and RSRES, highlighting the critical role of progressive cue modulation in accurate grounding and segmentation. 
Excluding CFM also degrades performance, indicating the importance of cross-scale feature fusion for capturing multi-scale information. 
Removing LCM primarily affects localization accuracy, with decreases of \textbf{4.06\%} and \textbf{3.26\%} in Pr@0.5 on RSREC and RSRES, respectively. 
Without the FA module in LCM, the performance drop is more pronounced on RSREC than on RSRES. 
This suggests that REC and RES rely on different information during decoding: REC requires more precise spatial localization, while RES depends more on semantic reasoning. 
The proposed FA module facilitates more effective multi-scale semantic integration, thereby improving grounding accuracy.

\begin{table}[t]
\centering
\caption{Ablation study on the weighting coefficients.}
\begin{tabular}{l|cccccc}
\toprule
        & \multicolumn{3}{c|}{RSREC} & \multicolumn{3}{c}{RSRES}\\
\cmidrule(r){2-7}
Exp. & Pr@0.5 & oIoU & mIoU & Pr@0.5 & oIoU & mIoU\\
\midrule
\multicolumn{7}{c}{(a) varying $\lambda_2$ with $\lambda_1$ and $\lambda_3$ fixed} \\
\midrule
$\lambda_2$=0.3 & 75.12 & 76.15 & 65.89 & 75.10 & 76.89 & 64.78   \\
$\lambda_2$=0.5 & \textbf{76.21} & \textbf{78.28} & \textbf{66.17} & \textbf{76.10} & \textbf{77.62} & \textbf{65.44}   \\
$\lambda_2$=0.7 & 75.26 & 75.94 & 65.78 & 75.56 & 77.01 & 64.89   \\
\midrule
\multicolumn{7}{c}{(b) varying $\lambda_3$ with $\lambda_1$ and $\lambda_2$ fixed} \\
\midrule
$\lambda_3$=0   & 74.82 & 75.30 & 65.03 & 75.82 & 77.08 & 64.54   \\
$\lambda_3$=0.1 & \textbf{76.21} & \textbf{78.28} & \textbf{66.17} & \textbf{76.10} & \textbf{77.62} & \textbf{65.44}   \\
$\lambda_3$=0.2 & 76.08 & 76.54 & 65.88 & 76.03 & 77.12 & 65.17   \\
\bottomrule
\end{tabular}
\label{tab:ablation2}
\end{table}

\noindent \textbf{Effect of weighting coefficients.}
Tab.~\ref{tab:ablation2} studies the impact of different weighting coefficients by varying $\lambda_2$ and $\lambda_3$ while keeping the remaining terms fixed. 
The best performance is achieved at $\lambda_2{=}0.5$ and $\lambda_3{=}0.1$, while deviations from this setting consistently degrade performance on both tasks. 
Notably, when $\lambda_3$ is set to 0, RSREC suffers the largest drop, indicating that this term is particularly important for accurate grounding. 
This observation is consistent with Fig.~\ref{fig:5}. \textbf{Q4}, where removing this constraint leads to misalignment between localization and segmentation.

\section{Conclusion}
\label{sec5}
This paper revisits remote sensing visual grounding (RSVG) from the perspective of structured perception and presents ProVG, a framework for progressive vision-language alignment. 
Inspired by psycholinguistic and neurocognitive evidence that human object identification follows a staged perception process, ProVG sequentially incorporates global context, spatial relations, and object attributes through a \emph{survey-locate-verify} scheme. 
By integrating progressive modulation, cross-scale fusion, and language-guided calibration, ProVG provides a unified multi-stage solution for both RSREC and RSRES. 
Extensive experiments on RRSIS-D and RISBench demonstrate the effectiveness of ProVG, particularly in complex scenarios with dense objects and ambiguous expressions.

Future work will explore more general progressive grounding strategies for large-scale vision-language models and extend ProVG to broader multimodal understanding tasks beyond remote sensing.

\bibliographystyle{ACM-Reference-Format}
\bibliography{sample-base}

\end{document}